\pdfoutput=1

\documentclass[11pt]{article}

\usepackage{acl}

\usepackage{times}
\usepackage{latexsym}

\usepackage[T1]{fontenc}

\usepackage[utf8]{inputenc}

\usepackage{microtype}

\usepackage{amsmath,amsthm,amsfonts,amssymb,amscd}
\usepackage{mathrsfs}
\usepackage{times}
\usepackage{graphicx} 
\usepackage{wrapfig}
\usepackage{subcaption}
\usepackage{multirow}
\usepackage{arydshln}

\title{Sharing Parameter by Conjugation for Knowledge Graph Embeddings \\ in Complex Space}

\author{
  Xincan Feng\textsuperscript{\dag \ddag}, Zhi Qu\textsuperscript{\dag}, Yuchang Cheng\textsuperscript{\ddag}, Taro Watanabe\textsuperscript{\dag}, Nobuhiro Yugami\textsuperscript{\ddag} \\
  \textsuperscript{\dag}Natural Language Processing Laboratory, Nara Institute of Science and Technology \\
  \textsuperscript{\ddag}Multilingual Knowledge Computing Laboratory, Fujitsu Ltd. \\
  \texttt{\{feng.xincan.fy2, qu.zhi.pv5, taro\}@is.naist.jp} \\
  \texttt{\{cheng.yuchang, yugami\}@fujitsu.com}\\
  }

\begin{document}
\maketitle
\begin{abstract}
A Knowledge Graph (KG) is the directed graphical representation of entities and relations in the real world. KG can be applied in diverse Natural Language Processing (NLP) tasks where knowledge is required. The need to scale up and complete KG automatically yields Knowledge Graph Embedding (KGE), a shallow machine learning model that is suffering from memory and training time consumption issues. To mitigate the computational load, we propose a parameter-sharing method, i.e., using conjugate parameters for complex numbers employed in KGE models. Our method improves memory efficiency by 2x in relation embedding while achieving comparable performance to the state-of-the-art non-conjugate models, with faster, or at least comparable, training time. We demonstrated the generalizability of our method on two best-performing KGE models $5^{\bigstar}\mathrm{E}$ \cite{Nay21} and $\mathrm{ComplEx}$ \cite{Tro16} on five benchmark datasets.
\end{abstract}

\section{Introduction}
A Knowledge Graph (KG) is a representation of confident information in the real world and employed in diverse Natural Language Processing (NLP) applications, e.g., recommender system, question answering, and text generation. A triple in the form of $(head, relation, tail)$ is widely used as the representation of elements in the KG instead of raw text for scalability. Cite $(clinician, synset\_domain\_topic\_of, psychology)$ as an example, $clinician$ and $psychology$ is the head and tail entity respectively, and $synset\_domain\_topic\_of$ is the relation of the head entity pointing to the tail entity. 

Knowledge Graph Embedding (KGE) models are designed for automatic link prediction. Relations in KG have multiple categories, e.g., symmetry, antisymmetry, inversion, and hierarchical. Missing links indicate incomplete ties between entities and are a common phenomenon as finding the missed connections is labor-intensive work. 

The theoretical space complexity of KGE models are often $\mathcal{O}(n_e d_e+n_r d_r)$, which is proportional to the number of KG elements, i.e. entities $n_e$ and relations $n_r$, and embedding dimension $d_e,d_r$ respectively. Scaling a KG is problematic as $n_e,n_r$ can go up to millions; also because KGE models are often shallow machine learning models composed of simple operations, e.g., matrix multiplication. Caution that a shallow model needs a large dimension size $d$ to depict the data feature, yielding the issue of the drastic increase of embedding parameters \cite{Det18}. 

KGE models represented using complex numbers have state-of-the-art performance, while they demand high memory costs. E.g., if using one of the best models $\mathrm{ComplEx}$ \cite{Tro16} to create embedding for the benchmark dataset FB15K whose $n_e=14,951,n_r=1,345$, and the best-performing dimensionalities $d_e=4000,d_r=4000$, will result in the parameter size of $65,184,000$. Considering the data type 64-bit integer (signed), who has a size of 8 bytes in PyTorch, the memory cost will be $65,184,000\times 8\approx 497$ MB. A KG for real-world application could have a much larger size, e.g., IBM's KG contains entities > 100 million and relations > 5 billion, which is actively in use and continually growing \cite{industry-kg}, would need > $148$ TB memory to do link prediction task.

Inspired by the improved performance of complex number representation and Non-Euclidean models where transformation parameters attempt to interact rather than be independent, we intuited the idea of sharing parameters for memory efficiency. 


We demonstrate a parameter-sharing method for complex numbers employed in KGE models. Specifically, our method formulates conjugate parameters in appropriate dimensions of the transformation functions to reduce relation parameters. By using our method, models can reduce their space complexity to $\mathcal{O}(n_e d_e+n_r d_r/2)$, which means the relation embedding size is half the original model. In the second place, using conjugate parameters may help save training time, especially on the datasets who have more parameter patterns. Further, our method can be easily applied to various complex number represented models. 

We verified our method on two best-performing KGE models, i.e., $\mathrm{ComplEx}$ \cite{Tro16} and $5^{\bigstar}\mathrm{E}$ \cite{Nay21}. The experiments were conducted on five benchmark datasets, i.e., FB15K-237, WN18RR, YAGO3-10, FB15K, and WN18, by which we empirically show that our method reserves the models' ability to achieve state-of-the-art results. We also see 31\% training time saved on average for $5^{\bigstar}\mathrm{E}$ in addition to the memory. Our method is implemented in PyTorch\footnote{\url{https://pytorch.org/}} and the code with hyperparameter settings\footnote{\url{github.com/xincanfeng/dimension}} are available online.

\section{Related Works}
We describe the categorizations of KGE models according to the representation method and the vector space that inspired our idea.

\paragraph{Representation Method}
Real and complex number representations are used to quantify entities and relations.

Translation approaches including $\mathrm{TransE}$ \cite{Bor13} and its variants \cite{Jie15, Lin15} describe embeddings using real number representation. Although these simple models cost fewer parameters, they can only encode two or three relation patterns, e.g., $\mathrm{TransE}$ cannot encode symmetric relations.

$\mathrm{ComplEx}$ \cite{Tro16} creates embedding with complex number representation, which can handle a wider variety of relations than using only real numbers, among them symmetric and antisymmetric relations \cite{Tro16}. $5^{\bigstar}\mathrm{E}$ \cite{Nay21} utilizes Möbius transformation, a projective geometric function that supports multiple simultaneous transformations in complex number representation and can embed entities in much lower ranks.

\paragraph{Vector Space}
Euclidean and Non-Euclidean spaces are practiced for the calculation of triple plausibility.

Factorization models such as $\mathrm{RESCAL}$ \cite{Nic11} and $\mathrm{DistMult}$ \cite{Yan15} employ element-wise multiplication in Euclidean space. Correspondingly, the plausibility of a triple is measured according to the angle of transformed head and tail entities.

$\mathrm{MuRP}$ \cite{Bal19} minimizes hyperbolic distances other than Euclidean. It needs fewer parameters than its Euclidean analog. $\mathrm{ATTH}$ \cite{Cha20} leverages trainable hyperbolic curvatures for each relation to simultaneously capture logical patterns and hierarchies. Compared with Euclidean, the Hyperbolic models can save more structures using variational curvatures in different areas to depict hierarchical relations.

\paragraph{Relational Constrain on Parameters}

Replacing real number with complex number representation enables the imaginary part to have an effect on the real part parameters, the boosted performance of which indicates the \textbf{hidden relation} among parameter. Using hyperbolic space other than Euclidean enables the distances or angles at different positions to vary, the increased accuracy hints us to add \textbf{various constraints} on parameters. Learning from the work by \citet{hayashi-shimbo-2017-equivalence}, the potential of improving representations through conjugate symmetric constraint is revealed. Therefore, we hypothesize the efficiency of relational parameters and propose a parameter-sharing method using conjugate numbers.

\section{Method}
Complex number employed in current KGE models enforces \textbf{multiplicative constraint} on representations; our method further adds \textbf{conjugate constraint} within the parameters. Note that we don't reduce the dimensions of the parameters, instead, we share the dimensions.

We economize 50\% of the memory in relation embedding by sharing half of the parameters in the conjugate form. Our approach is at least comparable in accuracy to the baselines. In addition, our method reduces calculation in the regularization process, e.g., for the $5^{\bigstar}\mathrm{\epsilon}$ model, 31\% of training time is saved on average for five benchmark datasets.

\subsection{Preliminaries}
Link prediction task inquires if a triple $(h,r,t)$ constructed by existing head and tail entities $h,t \in \mathbb{V}^{d_e}$ and relations $r \in \mathbb{V}^{d_r}$ ($\mathbb{V}^{d}$ is a $d$-dimensional vector space) is true or not. In KGE models, the relations are often represented as the transformation function $\vartheta$ that maps a head entity into a tail entity which are described as vectors in corresponding space, i.e., $\vartheta(h)=t$. Then, the score function $f: \mathbb{V}^{d_e} \times \mathbb{V}^{d_r} \times \mathbb{V}^{d_e} \rightarrow \mathbb{R}$ returns the plausibility $p$ of constructing a true triple: $f(h,r,t)=p(\vartheta(h),t)$. 

$a,b,c,d \in \mathbb{C}$ denote the parameters in the relation embedding matrices. $x \in \mathbb{C}$ is the parameter of the entity embedding matrices. $a_i,x_i$ are the parameters of the submatrices of $[a]$ and $[x]$ respectively. $Re(z)$ is the real part of the complex number $z$, $\overline{z}$ is the complex conjugate of $z$.

\paragraph{$\mathrm{\mathbf{ComplEx}}$}
This is the first and one of the best-performing complex models in Euclidean space. \citet{Tro16} demonstrated that complex number multiplication could capture antisymmetric relations while retaining the efficiency of the dot product, i.e., linearity in both space and time complexity. Balancing between model expressiveness and parameter size is also discussed as the keystone of KGE. However, targeting SOTA is still computational-expensive because \citet{Tro16} didn't solve the performance deterioration problem when reducing parameters directly. 

Performance deterioration can be severe whenever the KG needs to be expanded because the mispredicted links could lead to further misinformation. Hence we should always endeavour to adopt the best-performing embedding size in doing link prediction task, even though it could be hundreds of TB.

To obtain the best results, $\mathrm{ComplEx}$ needs embedding size of $rank=2000$ on dataset FB15K-237, WN18RR, FB15K, WN18, and $rank=1000$ on dataset YAGO3-10. $rank$ denotes the vector dimension of a single-functional parameter. Each entity and relation in this model needs $2\times rank$ parameters, representing real and imaginary part, respectively.

In this model, relations are represented as the real part of low-rank matrix
$\begin{bmatrix}
    a\\
\end{bmatrix}$, which act as weights on each entity dimension $x$, followed by a projection onto the real subspace. The transformation of $\mathrm{ComplEx}$ is
\begin{equation}
    x \rightarrow
    \begin{bmatrix}
        a
    \end{bmatrix}
    x
    \rightarrow
    ax.
\end{equation}

\paragraph{$\mathbf{5^{\bigstar}}\mathrm{\mathbf{E}}$}
This is a novel model applying complex numbers in Non-Euclidean space. \citet{Nay21} tackled the problem of multiple subgraph structures in the neighborhood, e.g., combinations of path and loop structures. Unlike the $\mathrm{ComplEx}$ model, they replaced the dot product with the Möbius function which has several favorable theoretical properties. This model subsumes $\mathrm{ComplEx}$ in that it embeds entities in much lower ranks, i.e., about 25\% or even smaller to achieve the state-of-the-art performance. However, $5^{\bigstar}\mathrm{E}$ is inferior to $\mathrm{ComplEx}$ in that it needs almost the same large size of relation parameters to do much more sophisticated calculation.

Following the hyperparameter search range of \citet{Nay21}, the embedding sizes we tested for $5^{\bigstar}\mathrm{E}$ to obtain the best result are $rank=500$ for all datasets. Each entity needs $2\times rank$ parameters, and each relation needs $8\times rank$ parameters that function differently.

In this model, relations are represented as 
$\begin{bmatrix}
    a & b\\
    c & d\\
\end{bmatrix}$. The transformation function $\vartheta$ of $5^{\bigstar}\mathrm{E}$ is
\begin{equation}
    x \rightarrow
    \begin{bmatrix}
        x\\
        1
    \end{bmatrix}
    \rightarrow
    \begin{bmatrix}
        a&b\\
        c&d\\
    \end{bmatrix}
    \begin{bmatrix}
        x\\
        1
    \end{bmatrix}
    \rightarrow
    \frac{ax+b}{cx+d}.
\end{equation}

Möbius function $\vartheta$ is capable of representing various relations simultaneously because it combines five subsequent transformations:
$\vartheta=\vartheta_4 \circ \vartheta_3 \circ \vartheta_2 \circ \vartheta_1$, where
$\vartheta_1=x+\frac{d}{c}$ is describing translation by $\frac{d}{c}$,
$\vartheta_2=\frac{1}{x}$ is describing inversion and reflection w.r.t. real axis,
$\vartheta_3=\frac{bc-ad}{c^2} x$ is describing homothety and rotation, and
$\vartheta_4=x+\frac{a}{c}$ is describing translation by $\frac{a}{c}$.

\begin{figure*}[htbp!]
    \centering
    \begin{subfigure}{0.49\textwidth}
        \vspace{0.5cm}
        \includegraphics[scale=0.05]{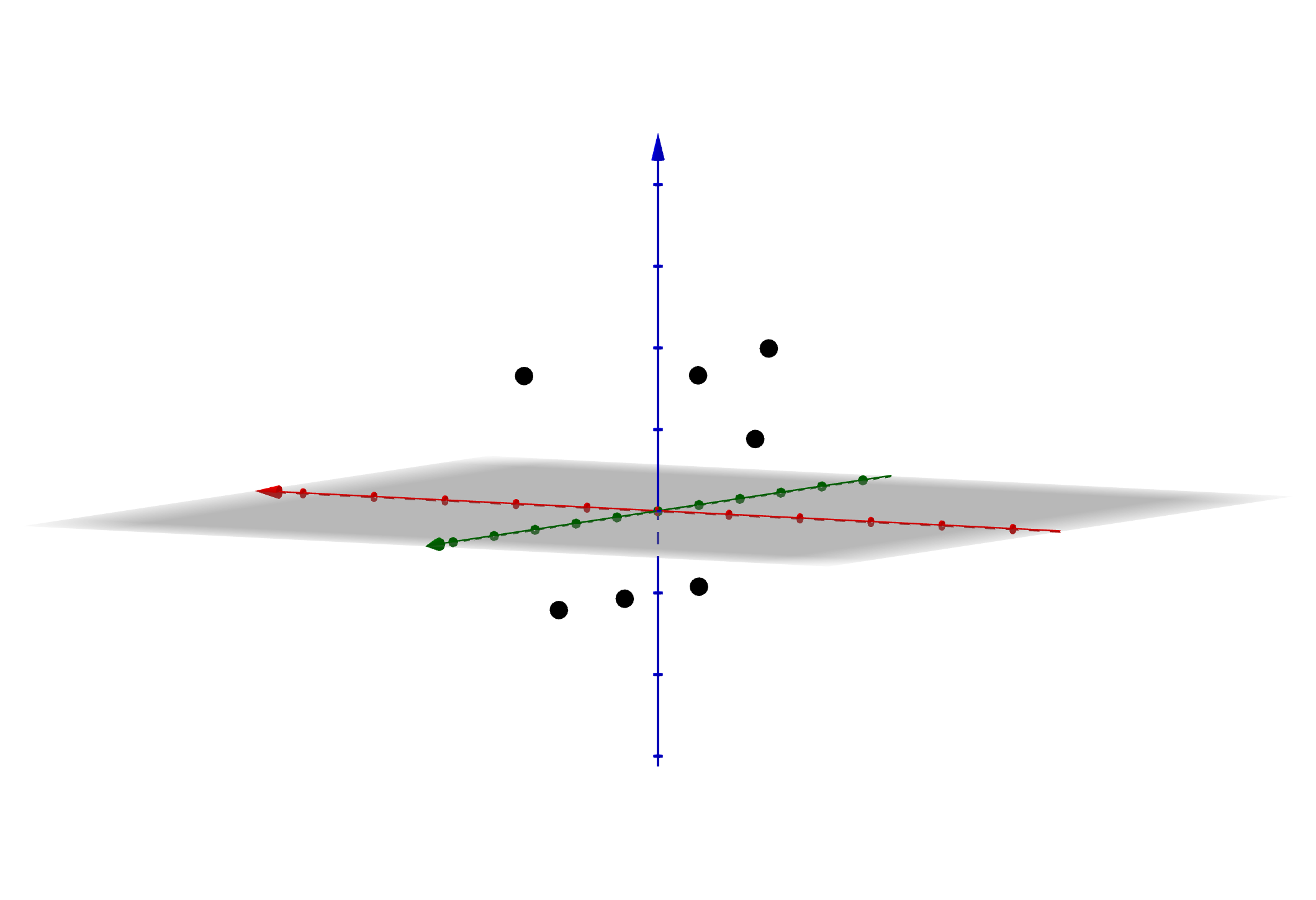}
        \includegraphics[scale=0.05]{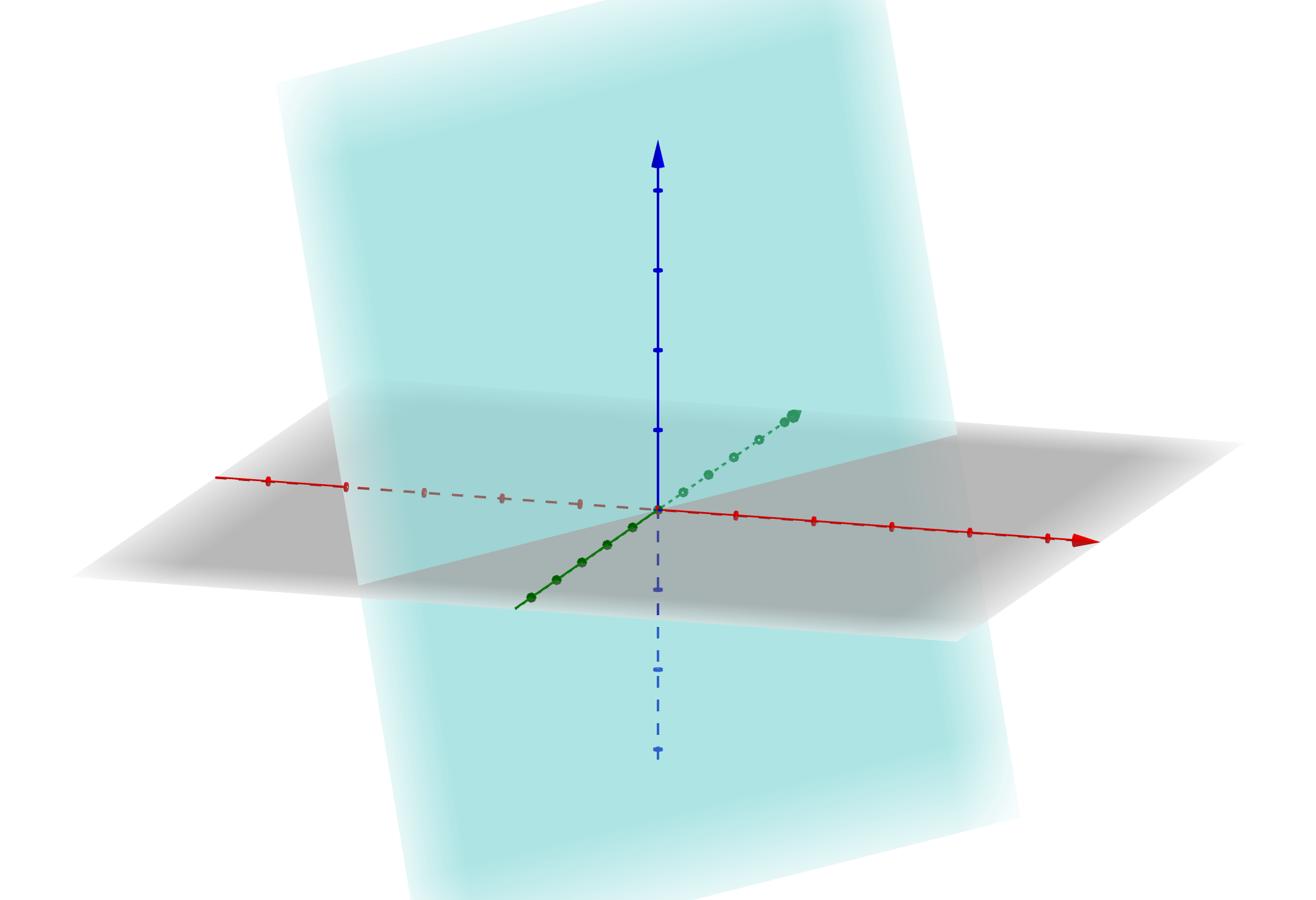}
        \vspace{0.5cm}
        \caption{
            Transformed entities by $\mathrm{ComplEx}$ (left) and $\mathrm{Compl\epsilon x}$ (right). In the left graph, a black point describes a transformed entity, and the vector values of a point are unrelated in each dimension. 
            While in the right graph, half of the value $z_i, i\in [1,d/2]$ of a vector $z_1,z_2,...,z_d$ that is describing a point are constrained as the other half $z_i, i\in [d/2+1,d]$ correspondingly. The linear constrain $z_i=a_ix_i+b_iy_i$ is illustrated in the right graph.}
    \end{subfigure}\hfill
    \begin{subfigure}{0.49\textwidth}
        \vspace{0.15cm}
        \includegraphics[scale=0.15]{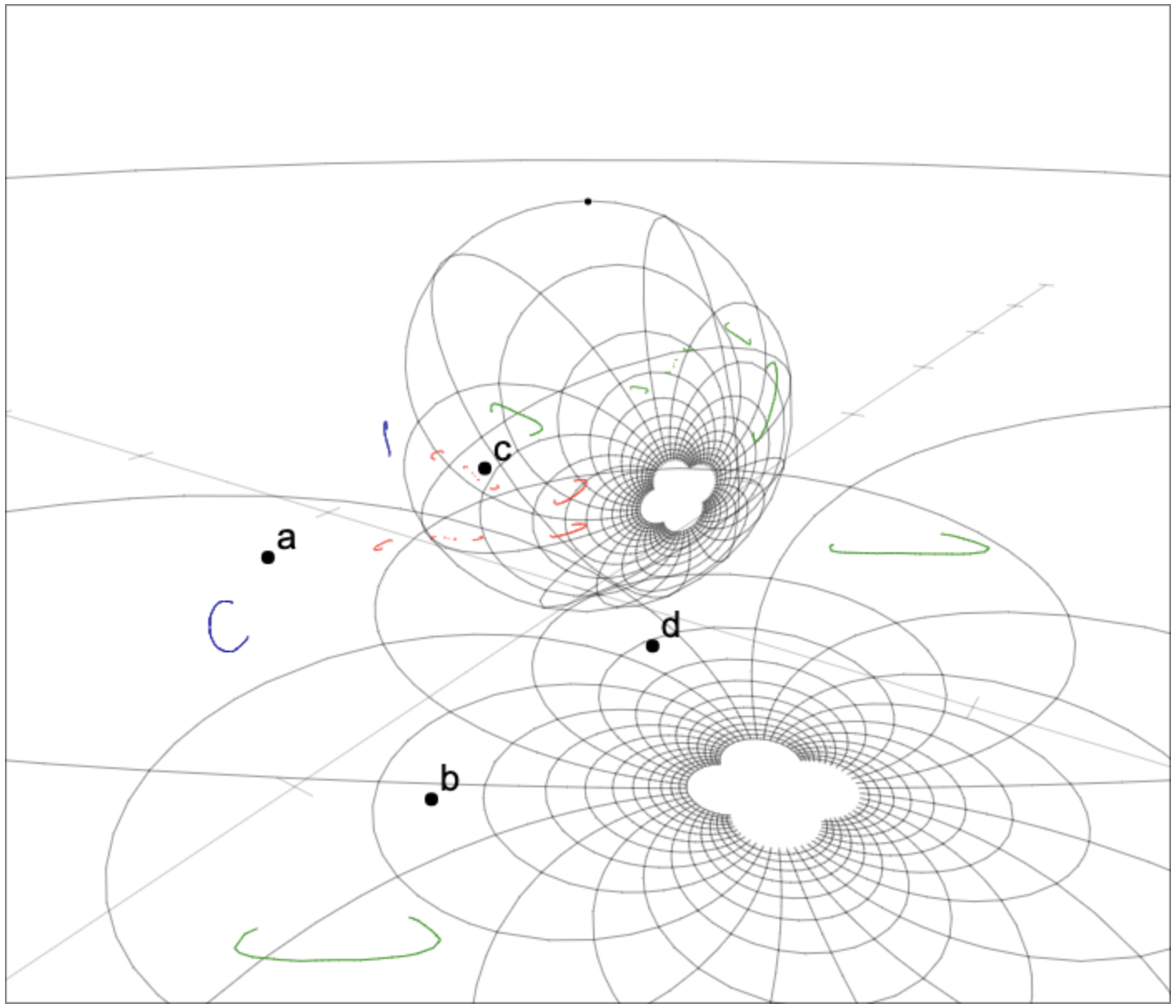}
        \includegraphics[scale=0.15]{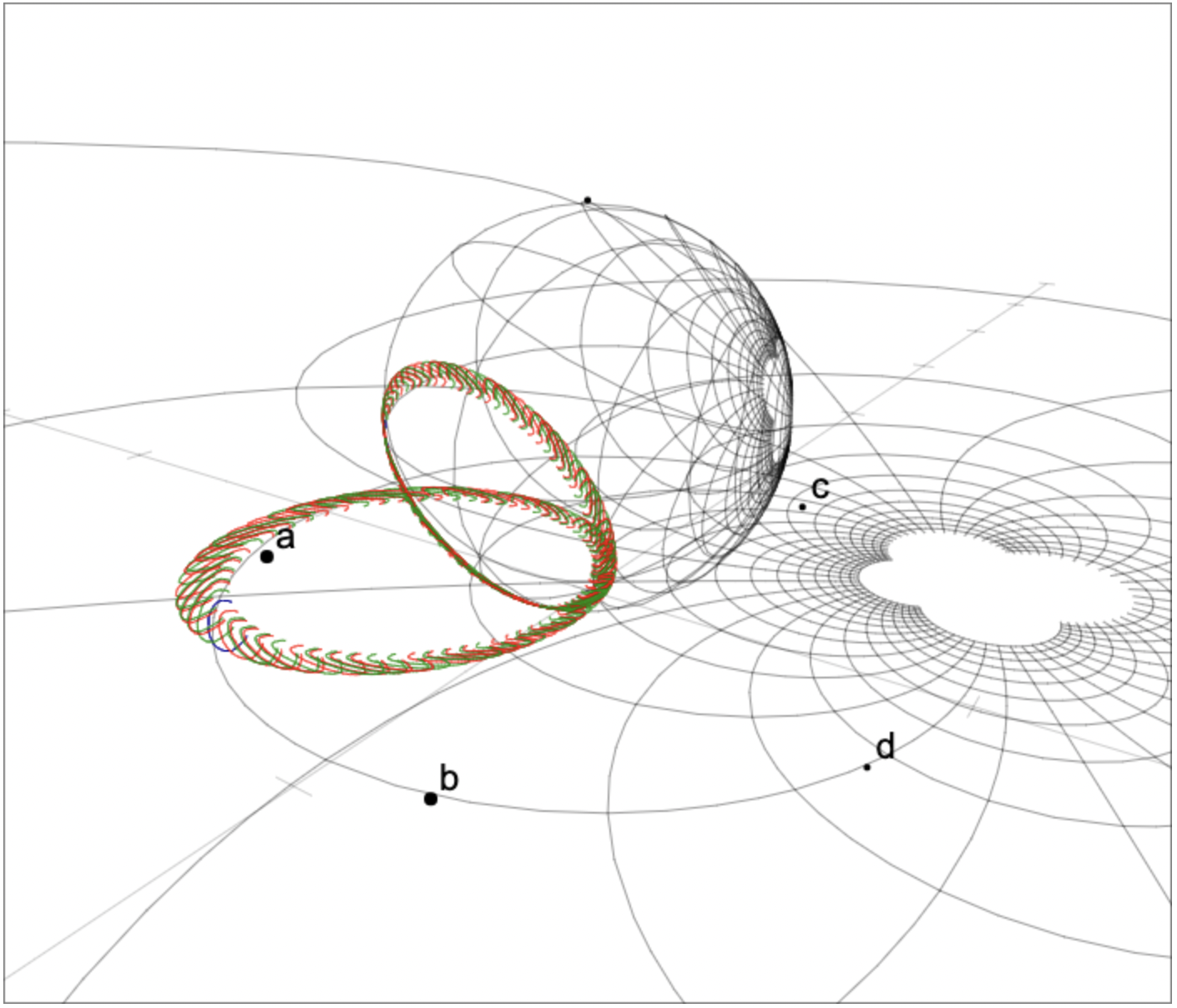}
        \vspace{0.15cm}
        \caption{
            Transformed entities by $5^{\bigstar}\mathrm{E}$ (left) and $5^{\bigstar}\mathrm{\epsilon}_n$ (right). Note that we illustrate the negative conjugated model instead of the positive conjugated one for simplicity in plotting. Blue traces are the original entities and their projections in the Non-Euclidean space. Green traces are the multiple copies of the blue traces under iterations of the Möbius transformation. Red traces are the inverse of green traces. Apparently, the right graph has much neater geometric properties.}
    \end{subfigure}
    \caption{Transformed entities illustrated in 3D}
    \label{entities}
\end{figure*}

\subsection{Method Formulation}
Let $\begin{bmatrix}
    a_1 & a_2\\
\end{bmatrix}$ 
denotes the relation embedding matrix. Our method constrains half of the parameters $a_2$ using the complex conjugate of the other half $\overline{a_1}$, i.e., $a_2 = \overline{a_1}$; it is model-dependent to specify which parameters are suitable for conjugation. We formulated our method on above two baseline models.

\paragraph{$\mathrm{\mathbf{Compl\epsilon x}}$}
By using our method, the original model $\mathrm{ComplEx}$ is adapted to the parameter-sharing model $\mathrm{Compl\epsilon x}$, where relations are represented as the real part of low-rank matrices with conjugate parameters. Specifically, we set the original square relation embedding matrices 
$\begin{bmatrix}
    a_1 & a_2\\
\end{bmatrix}$
to be half the normal parameters and the other half their conjugation, i.e.,
$\begin{bmatrix}
    a_1 & \overline{a_1}\\
\end{bmatrix}$.
In this model, since each parameter is functioning equally, the positions of the conjugate parameters can be set randomly.

\paragraph{$\mathbf{5^{\bigstar}}\mathrm{\mathbf{\epsilon}}$}
Our method transforms the original model $5^{\bigstar}\mathrm{E}$ into the parameter-sharing model $5^{\bigstar}\mathrm{\epsilon}$, where relations are represented as the real part of low-rank matrices using conjugate parameters. Specifically, we set the original square relation embedding matrices 
$\begin{bmatrix}
    a & b\\
    c & d\\
\end{bmatrix}$
to be half the normal parameters and the other half their conjugate parameters, i.e.,
$\begin{bmatrix}
    a & b\\
    \overline{b} & \overline{a}\\
\end{bmatrix}$.
In this model, parameters play distinct roles at different positions, and the best conjugation positions are the principal and secondary diagonal positions. Note that experiments showed that, the following negative conjugation method, i.e.,
$\begin{bmatrix}
    a & b\\
    -\overline{b} & \overline{a}\\
\end{bmatrix}$,
achieves similar performance as above. Although the negative conjugation on this model is equivalent to restricting the original Möbius function to the unitary Möbius transformation, our approach is much more general to a variety of representations.

\subsection{Transformation Analysis}
\paragraph{$\mathrm{\mathbf{Compl\epsilon x}}$}
Let $a_2=\overline{a_1}$, then the transformation of conjugate model $\mathrm{Compl\epsilon x}$ is
\begin{equation}
    \begin{bmatrix}
        x_1 & x_2
    \end{bmatrix} 
    \rightarrow
    \begin{bmatrix}
        a_1 & \overline{a_1}
    \end{bmatrix}
    \begin{bmatrix}
        x_1 & x_2
    \end{bmatrix}
    \rightarrow
    \begin{bmatrix}
        a_1x_1 & \overline{a_1}x_2
    \end{bmatrix}.
\end{equation}
We can see that the resulted relation embedding is constrained to
$\begin{bmatrix}
    a_1 & \overline{a_1}
\end{bmatrix}$ other than
$\begin{bmatrix}
    a_1 & a_2
\end{bmatrix}$; the predicted tail entity is constrained to
$\begin{bmatrix}
    a_1x_1 & \overline{a_1}x_2
\end{bmatrix}$ instead of 
$\begin{bmatrix}
    a_1x_1 & a_2x_2
\end{bmatrix}$ in original model, which does not narrow the rang of relation or tail embedding since the $a_1, x_2$ can be any value. Further, since tail entities also act as head entities, we can say that the range of both the entities and relations are not constrained.

\begin{table*}[htbp]								
\centering								
\begin{tabular}{lrrrrrr}								
	\hline
        {Dataset}	&{\#Train}	&{\#Valid}	&{\#Test}	&{Ent}	&{Rel} &{Exa}\\
        \hline
	FB15K-237	&272,115	&17,535	&20,466	&14,541	&237	&544,230	\\
	WN18RR	&86,835	&3,034	&3,134	&40,943	&11	&173,670	\\
	YAGO3-10	&1,079,040	&5,000	&5,000	&123,188	&37	&2,158,080	\\
	FB15K	&483,142	&50,000	&59,071	&14,951	&1,345	&966,284	\\
	WN18	&141,442	&5,000	&5,000	&40,943	&18	&282,884	\\
	\hline
\end{tabular}								
\caption{Datasets statistics. \#: Split in terms of number of triples; Ent: Entities; Rel: Relations; Exa: Examples.}								
\label{datasets}								
\end{table*}								

\paragraph{$\mathbf{5^{\bigstar}}\mathrm{\mathbf{\epsilon}}$}
Let $c = \overline{b}, d = \overline{a}$, then the transformation of conjugate model $5^{\bigstar}\mathrm{\epsilon}$ is
\begin{equation}
    x \rightarrow
    \begin{bmatrix}
        x\\
        1
    \end{bmatrix}
    \rightarrow
    \begin{bmatrix}
        a & b\\
        \overline{b} & \overline{a}\\
    \end{bmatrix}
    \begin{bmatrix}
        x\\
        1
    \end{bmatrix}
    \rightarrow
    \frac{ax+b}{\overline{b}x+\overline{a}}. 
\end{equation}

The five subsequent transformations turn into:
$\vartheta_1=x+\frac{\overline{a}}{\overline{b}}$ which depicts translation by $\frac{\overline{a}}{\overline{b}}$,
$\vartheta_2=\frac{1}{x}$ which depicts inversion and reflection w.r.t. real axis,
$\vartheta_3=\frac{b\overline{b}-a\overline{a}}{\overline{b}^2} x$ which depicts homothety and rotation, and
$\vartheta_4=x+\frac{a}{\overline{b}}$ which depicts translation by $\frac{a}{\overline{b}}$.
We can see that, although the relation parameters are constrained comparing to the original model, the five sub-transformations are reserved in this conjugate model.

\paragraph{Characteristics}
For this reason, we consider our conjugate models retain expressiveness in function level for various relation patterns compared to their original counterparts. The difference between original models and our conjugate models is that, the latter ones have more linear constrain in its value of each embedding parameter, as illustrated in Figure \ref{entities}.

\subsection{Reduced Calculation}
Sharing half of the parameters also reduces the computation for the regularization terms into half, where each parameter of relation is squared to the sum. For example, the original calculation $r_1^2+r_2^2$ is turned into $r_1^2 \times 2$ in both baseline models, where $r_1,r_2$ denote the real or imaginary part of a complex number, and in which $r_1$ represents the shared parameter. However, the final time consumption depends on multiple aspects, such as formulation and coding, thus is not necessarily reduced.

\section{Experiments}
\subsection{Experimental Setup}
\paragraph{Metrics} 
We followed the standard evaluation protocal for KGE models. $T$: the rank set of truth, $r_i$: the rank position $r$ of the first true entity for the $i$-th query. We computed two rank-based metrics: (i) Mean Reciprocal Rank (MRR), which computes the arithmetic mean of reciprocal ranks of all true entities from the ranked list of answers to queries $T$, and (ii) Hits@$N$ ($N$ = 1, 3, 10), which counts the true entities $\mathbb{I}$ and calculate their proportion in the truth $T$ in top $N$ sorted predicted answers list.
\begin{equation}
    \mathrm{MRR}=\frac{1}{T}\sum_{i=1}^{T}\frac{1}{r_i}
\end{equation}
\begin{equation}
    \mathrm{Hits@}N=\frac{1}{T}\sum_{r\in T, r\leq N} \mathbb{I}
\end{equation}
We also use additional metric Time (seconds/epoch) to measure how many seconds each training epoch costs to demonstrate the time saved by our method. To do this, we conducted all experiments using the same GPUs. GeForce GTX 1080 Ti is used for all datasets except for the largest dataset YAGO3-10 who needs a larger GPU and we used Tesla V100S-PCIE-32GB for it. 

\paragraph{Datasets}
We evaluated our method on five widely used benchmark datasets (See Table \ref{datasets}). FB15K \cite{Bor13} is a subset of Freebase, the contents of which are general facts. WN18 \cite{Bor13} is a subset of Wordnet, a database that features lexical relations between words. YAGO3-10 \cite{Det18} is the largest common dataset, which mostly describes attributes of persons, and contains entities associated with at least ten different relations.

As was first noted by \citet{Tou15}, FB15K and WN18 suffer from test leakage through inverse relations, e.g., the test set frequently contains triples such as $(s, hyponym, o)$ while the training set contains its inverse $(o, hypernym, s)$. To create a dataset without this property, they introduced FB15K-237, a subset of FB15K where inverse relations are removed. WN18RR was created for the same reason by \citet{Det18}. 

We adopted all of the five datasets for comprehensive comparison of models.

\paragraph{Hyperparameter Settings}

We explored the influence of hyperparameter settings to our method. To do this, we used the best hyperparameter settings for the original models (marked as $\nabla$ or no mark), and applied the same settings on our conjugate models and ablation models. We adopted the best hyperparameter settings for $\mathrm{ComplEx}$ provided by \citet{Nay21}, and fine-tuned the best hyperparameters ourselves for $5^{\bigstar}\mathrm{E}$ since there was no published best hyperparameter settings for this model at the time we did the experiments. We also fine-tuned the best hyperparameters for one of our conjugate model $5^{\bigstar}\mathrm{\epsilon}$ (noted as $\diamondsuit$) to explore the upper bound.

We selected the hyperparameters based on the MRR on the validation set. Our grid search range refered to but was larger than \citet{Nay21}. The optional optimizers are \{Adagrad, Adam, SGD\}. The range of embedding dimensions are \{100, 500\} with learning rates range in \{1E-02, 5E-02, 1E-01\}. The batch sizes attempted range in \{100, 500, 1000, 2000\}. Regularization coefficients are tested among \{2.5E-03, 5E-03, 1E-02, 5E-02, 1E-01, 5E-01\}.

\begin{table}[t]												
\begin{subtable}[htbp]{.5\textwidth}											
\centering	

\resizebox{1\textwidth}{!}{										
\begin{tabular}{llllll}												
	\hline											
	Model	&	Time	&	MRR	&	H@1	&	H@3	&	H@10	\\
	\hline											
	$\mathrm{ComplEx}$	&	42$\pm$8	&	\textbf{0.366$\pm$4e-4}	&	\textbf{0.271}	&	\textbf{0.402}	&	\textbf{0.558}	\\
	$\mathrm{Compl\epsilon x}$	&	46$\pm$11	&	0.363$\pm$5e-4	&	0.268	&	0.400	&	0.555	\\
	$5^{\bigstar}\mathrm{E}$	&	18$\pm$3	&	0.350$\pm$8e-4	&	0.257	&	0.386	&	0.538	\\
	$5^{\bigstar}\mathrm{\epsilon}_\nabla$	&	\textbf{14$\pm$4}	&	0.353$\pm$7e-4	&	0.259	&	0.390	&	0.541	\\
	$5^{\bigstar}\mathrm{\epsilon}_\diamondsuit$	&	\textbf{17$\pm$9}	&	0.354$\pm$8e-4	&	0.259	&	0.391	&	0.544	\\
	\hline											
\end{tabular}}												
\caption{FB15K-237}												
\end{subtable}												
\begin{subtable}[htbp]{.5\textwidth}												
\centering												
\resizebox{1\textwidth}{!}{												
\begin{tabular}{llllll}												
	\hline											
	Model	&	Time	&	MRR	&	H@1	&	H@3	&	H@10	\\
	\hline											
	$\mathrm{ComplEx}$	&	139$\pm$21	&	0.488$\pm$1e-3	&	0.442	&	0.503	&	0.579	\\
	$\mathrm{Compl\epsilon x}$	&	146$\pm$45	&	0.475$\pm$9e-4	&	0.433	&	0.488	&	0.558	\\
	$5^{\bigstar}\mathrm{E}$	&	16$\pm$1	&	0.490$\pm$5e-4	&	\textbf{0.444}	&	0.506	&	0.587	\\
	$5^{\bigstar}\mathrm{\epsilon}_\nabla$	&	\textbf{11$\pm$1}	&	\textbf{0.493$\pm$8e-4}	&	0.442	&	\textbf{0.512}	&	0.588	\\
	$5^{\bigstar}\mathrm{\epsilon}_\diamondsuit$	&	-	&	-	&	-	&	-	&	-	\\
	\hline											
\end{tabular}}												
\caption{WN18RR}												
\end{subtable}												
\begin{subtable}[htbp]{.5\textwidth}												
\centering												
\resizebox{1\textwidth}{!}{												
\begin{tabular}{llllll}												
	\hline											
	Model	&	Time	&	MRR	&	H@1	&	H@3	&	H@10	\\
	\hline											
	$\mathrm{ComplEx}$	&	370$\pm$2	&	\textbf{0.577$\pm$1e-3}	&	0.502	&	\textbf{0.622}	&	\textbf{0.712}	\\
	$\mathrm{Compl\epsilon x}$	&	371$\pm$2	&	0.574$\pm$2e-3	&	0.500	&	0.618	&	0.707	\\
	$5^{\bigstar}\mathrm{E}$	&	415$\pm$2	&	0.574$\pm$2e-3	&	0.502	&	0.617	&	0.701	\\
	$5^{\bigstar}\mathrm{\epsilon}_\nabla$	&	\textbf{297$\pm$1}	&	0.576$\pm$2e-3	&	\textbf{0.505}	&	0.619	&	0.702	\\
	$5^{\bigstar}\mathrm{\epsilon}_\diamondsuit$	&	-	&	-	&	-	&	-	&	-	\\
	\hline											
\end{tabular}}												
\caption{YAGO3-10}												
\end{subtable}												
\caption{Link prediction results on FB15K-237, WN18RR, YAGO3-10 datasets. Time, MRR and H@n are presented as mean ($\pm$ standard deviation).}												
\label{mrrs1}												
\end{table}												
												
\begin{table}[t]												
\begin{subtable}[htbp]{.5\textwidth}												
\centering												
\resizebox{1\textwidth}{!}{												
\begin{tabular}{llllll}												
	\hline											
	Model	&	Time	&	MRR	&	H@1	&	H@3	&	H@10	\\
	\hline											
	$\mathrm{ComplEx}$	&	346$\pm$124	&	\textbf{0.855$\pm$1e-3}	&	0.823	&	\textbf{0.874}	&	\textbf{0.910}	\\
	$\mathrm{Compl\epsilon x}$	&	293$\pm$16	&	\textbf{0.855$\pm$1e-3}	&	\textbf{0.827}	&	0.871	&	0.907	\\
	$5^{\bigstar}\mathrm{E}$	&	42$\pm$9	&	0.812$\pm$1e-3	&	0.767	&	0.840	&	0.889	\\
	$5^{\bigstar}\mathrm{\epsilon}_\nabla$	&	\textbf{26$\pm$0}	&	0.794$\pm$2e-3	&	0.743	&	0.827	&	0.882	\\
	$5^{\bigstar}\mathrm{\epsilon}_\diamondsuit$	&	29$\pm$5	&	0.813$\pm$2e-3	&	0.766	&	0.844	&	0.894	\\
	\hline											
\end{tabular}}												
\caption{FB15K}												
\end{subtable}												
\begin{subtable}[htbp]{.5\textwidth}												
\centering												
\resizebox{1\textwidth}{!}{												
\begin{tabular}{llllll}												
	\hline											
	Model	&	Time	&	MRR	&	H@1	&	H@3	&	H@10	\\
	\hline											
	$\mathrm{ComplEx}$	&	57$\pm$3	&	0.951$\pm$3e-4	&	0.944	&	0.954	&	0.961	\\
	$\mathrm{Compl\epsilon x}$	&	58$\pm$5	&	0.950$\pm$3e-4	&	0.945	&	0.953	&	0.960	\\
	$5^{\bigstar}\mathrm{E}$	&	43$\pm$6	&	\textbf{0.952$\pm$5e-4}	&	0.946	&	\textbf{0.955}	&	\textbf{0.962}	\\
	$5^{\bigstar}\mathrm{\epsilon}_\nabla$	&	\textbf{29$\pm$6}	&	0.949$\pm$6e-4	&	0.944	&	0.953	&	0.959	\\
	$5^{\bigstar}\mathrm{\epsilon}_\diamondsuit$	&	\textbf{26$\pm$2}	&	\textbf{0.952$\pm$3e-4}	&	\textbf{0.947}	&	\textbf{0.955}	&	\textbf{0.962}	\\
	\hline											
\end{tabular}}												
\caption{WN18}												
\end{subtable}												
\caption{Link prediction results on FB15K and WN18 datasets. Instructions for this table are the same as those in Table \ref{mrrs1}.}												
\label{mrrs2}												
\end{table}												
\begin{table}[t]												
\begin{subtable}[htbp]{.5\textwidth}												
\centering												
\resizebox{1\textwidth}{!}{												
\begin{tabular}{llllll}												
	\hline											
	Model	&	Time	&	MRR	&	H@1	&	H@3	&	H@10	\\
	\hline											
	$5^{\bigstar}\mathrm{\epsilon}_\nabla$	&	\textbf{14$\pm$4}	&	0.353$\pm$7e-4	&	0.259	&	0.390	&	0.541	\\
	$5^{\bigstar}\mathrm{\epsilon}_n$	&	\textbf{13$\pm$2}	&	0.353$\pm$8e-4	&	0.259	&	0.389	&	0.541	\\
	$5^{\bigstar}\mathrm{E_r}$	&	16$\pm$0	&	0.326$\pm$1e-3	&	0.238	&	0.357	&	0.505	\\
	$5^{\bigstar}\mathrm{\epsilon}_v$	&	13$\pm$1	&	0.264$\pm$4e-4	&	0.192	&	0.288	&	0.404	\\
	$5^{\bigstar}\mathrm{\epsilon}_h$	&	12$\pm$0	&	0.301$\pm$4e-4	&	0.221	&	0.329	&	0.458	\\
	\hline											
\end{tabular}}												
\caption{FB15K-237}												
\end{subtable}												
\begin{subtable}[htbp]{.5\textwidth}												
\centering												
\resizebox{1\textwidth}{!}{												
\begin{tabular}{llllll}												
	\hline											
	Model	&	Time	&	MRR	&	H@1	&	H@3	&	H@10	\\
	\hline											
	$5^{\bigstar}\mathrm{\epsilon}_\nabla$	&	\textbf{11$\pm$1}	&	\textbf{0.493$\pm$8e-4}	&	0.442	&	\textbf{0.512}	&	0.588	\\
	$5^{\bigstar}\mathrm{\epsilon}_n$	&	14$\pm$3	&	0.485$\pm$1e-3	&	0.432	&	0.506	&	\textbf{0.589}	\\
	$5^{\bigstar}\mathrm{E_r}$	&	16$\pm$0	&	0.410$\pm$3e-3	&	0.391	&	0.417	&	0.447	\\
	$5^{\bigstar}\mathrm{\epsilon}_v$	&	12$\pm$5	&	0.026$\pm$2e-4	&	0.015	&	0.025	&	0.045	\\
	$5^{\bigstar}\mathrm{\epsilon}_h$	&	14$\pm$3	&	0.026$\pm$3e-4	&	0.016	&	0.025	&	0.046	\\
	\hline											
\end{tabular}}												
\caption{WN18RR}												
\end{subtable}												
\begin{subtable}[htbp]{.5\textwidth}												
\centering												
\resizebox{1\textwidth}{!}{												
\begin{tabular}{llllll}												
	\hline											
	Model	&	Time	&	MRR	&	H@1	&	H@3	&	H@10	\\
	\hline											
	$5^{\bigstar}\mathrm{\epsilon}_\nabla$	&	\textbf{297$\pm$1}	&	0.576$\pm$2e-3	&	\textbf{0.505}	&	0.619	&	0.702	\\
	$5^{\bigstar}\mathrm{\epsilon}_n$	&	\textbf{298$\pm$1}	&	0.574$\pm$1e-3	&	0.502	&	0.618	&	0.701	\\
	$5^{\bigstar}\mathrm{E_r}$	&	416$\pm$2	&	0.569$\pm$2e-3	&	0.499	&	0.611	&	0.695	\\
	$5^{\bigstar}\mathrm{\epsilon}_v$	&	297$\pm$1	&	0.562$\pm$8e-4	&	0.488	&	0.607	&	0.695	\\
	$5^{\bigstar}\mathrm{\epsilon}_h$	&	298$\pm$1	&	0.546$\pm$1e-3	&	0.471	&	0.592	&	0.680	\\
	\hline											
\end{tabular}}												
\caption{YAGO3-10}												
\end{subtable}												
\caption{Ablation studies on FB15K-237, WN18RR, YAGO3-10 datasets. Instructions for this table are the same as those in Table \ref{mrrs1}.}												
\label{mrrs3}												
\end{table}												
												
\begin{table}[t]												
\begin{subtable}[htbp]{.5\textwidth}												
\centering												
\resizebox{1\textwidth}{!}{												
\begin{tabular}{llllll}												
	\hline											
	Model	&	Time	&	MRR	&	H@1	&	H@3	&	H@10	\\
	\hline											
	$5^{\bigstar}\mathrm{\epsilon}_\nabla$	&	\textbf{26$\pm$0}	&	0.794$\pm$2e-3	&	0.743	&	0.827	&	0.882	\\
	$5^{\bigstar}\mathrm{\epsilon}_n$	&	\textbf{31$\pm$12}	&	0.799$\pm$2e-3	&	0.750	&	0.831	&	0.883	\\
	$5^{\bigstar}\mathrm{E_r}$	&	37$\pm$1	&	0.807$\pm$3e-3	&	0.760	&	0.838	&	0.888	\\
	$5^{\bigstar}\mathrm{\epsilon}_v$	&	31$\pm$7	&	0.801$\pm$8e-4	&	0.753	&	0.833	&	0.885	\\
	$5^{\bigstar}\mathrm{\epsilon}_h$	&	28$\pm$2	&	0.787$\pm$2e-3	&	0.735	&	0.822	&	0.877	\\
	\hline											
\end{tabular}}												
\caption{FB15K}												
\end{subtable}												
\begin{subtable}[htbp]{.5\textwidth}												
\centering												
\resizebox{1\textwidth}{!}{												
\begin{tabular}{llllll}												
	\hline											
	Model	&	Time	&	MRR	&	H@1	&	H@3	&	H@10	\\
	\hline											
	$5^{\bigstar}\mathrm{\epsilon}_\nabla$	&	\textbf{29$\pm$6}	&	0.949$\pm$6e-4	&	0.944	&	0.953	&	0.959	\\
	$5^{\bigstar}\mathrm{\epsilon}_n$	&	\textbf{26$\pm$0}	&	\textbf{0.952$\pm$3e-4}	&	0.946	&	\textbf{0.955}	&	\textbf{0.962}	\\
	$5^{\bigstar}\mathrm{E_r}$	&	40$\pm$0	&	0.943$\pm$9e-4	&	0.935	&	0.950	&	0.954	\\
	$5^{\bigstar}\mathrm{\epsilon}_v$	&	31$\pm$11	&	0.892$\pm$2e-3	&	0.836	&	0.944	&	0.958	\\
	$5^{\bigstar}\mathrm{\epsilon}_h$	&	26$\pm$0	&	0.822$\pm$2e-3	&	0.719	&	0.920	&	0.949	\\
	\hline											
\end{tabular}}												
\caption{WN18}												
\end{subtable}												
\caption{Ablation studies on FB15K and WN18 datasets. Instructions for this table are the same as those in Table \ref{mrrs1}.}												
\label{mrrs4}												
\end{table}												

\section{Results}
\subsection{Main Results and Analysis}
The main experimental results are shown in Table \ref{mrrs1} and Table \ref{mrrs2}. The numbers with boldface indicate the best results among all the models.

We mainly tested whether the conjugate models perform consistent with their original counterparts, especially whether they can achieve the same state-of-the-art results. We conducted one set of experiments using the best hyperparameters of the original models (marked as $\nabla$ or no mark), and the other set of experiments tuning the hyperparameters for one of our conjugate model (marked as $\diamondsuit$). 

The results show that both $\mathrm{Compl\epsilon x}$ and $5^{\bigstar}\mathrm{\epsilon}$ consistently achieve results comparable to their original models on the datasets without test set leakage, including the largest dataset, i.e., YAGO3-10; and obtain the same optimal accuracies as the original models on all datasets with possibly-required fine-tuning. From the perspective of training time, we see $5^{\bigstar}\mathrm{\epsilon}$ spends 31\% less time on average for all datasets; and both conjugate models perform substantially best in training time on datasets FB15K, who have the most relations.

\paragraph{$\mathrm{\mathbf{Compl\epsilon x}}$}
Under the best hyperparameter settings of the original model, the performance of $\mathrm{Compl\epsilon x}$ are consistently comparable with $\mathrm{ComplEx}$ on all five datasets. We speculate the reason for the consistent but tiny performance drop might come from the computation precision, but we will leave it as our future studies.

Although our method reduces the computation, applying the method on this model requires splitting and concatenating matrices to keep the shape of outputs which incurs additional time-consuming operations. Consequently, the total time cost is not reduced much. However, training time on dataset FB15K, who has the most relations, becomes very stable.

Overall results imply our conjugate model $\mathrm{Compl\epsilon x}$ is at least comparative with its baseline model $\mathrm{ComplEx}$.

\paragraph{$\mathbf{5^{\bigstar}}\mathrm{\mathbf{\epsilon}}$}
Under the best hyperparameter settings of the original $5^{\bigstar}\mathrm{E}$, the conjugate $5^{\bigstar}\mathrm{\epsilon}_\nabla$ consistently achieve competitive results on the datasets FB15K-237, WN18RR and YAGO3-10. The tiny but consistent accuracy enhancement on these three datasets is probably caused by similar programming artifacts as observed in $\mathrm{ComplEx}$.

We hypothesize that the accuracy fluctuation of $5^{\bigstar}\mathrm{\epsilon}_\nabla$ on FB15K and WN18 is caused by the test leakage issue which makes the model sensitive to its hyperparameter setting. Because the only difference of these two datasets comparing to their subsets FB15K-237 and WN18RR is the 81\% and 94\% inverse relations \cite{Tou15}, i.e., $(s, hyponym, o)$ and $(o, hypernym, s)$ in the training set and the test set respectively, which is known as test leakage. Note that the accuracy fluctuation was simply solved by fine-tuning the hyperparameters (See results marked as $5^{\bigstar}\mathrm{\epsilon}_\diamondsuit$).

Notice that in Table \ref{mrrs1}, we didn't report the fine-tuned results of $5^{\bigstar}\mathrm{\epsilon}_\diamondsuit$ on datasets WN18RR and YAGO3-10, because the results abtained with the original settings $\nabla$ is already the best.

Training time in this model was reduced by 22\%, 31\%, 28\%, 38\% and 33\% on each dataset respectively, and 31\% on average. $5^{\bigstar}\mathrm{E}$ has eight parameter matrices in the coding. By using our method, the parameter matrices are directly reduced to four with no additional coding operations, which makes the significant saved training time.

Above results mean our conjugate model $5^{\bigstar}\mathrm{\epsilon}$ exceeds the baseline model $5^{\bigstar}\mathrm{E}$ in all respect of accuracy, memory-efficiency and time footprint.

\subsection{Ablation Studies}
We did two kinds of ablation studies. The results are shown in Table \ref{mrrs3} and Table \ref{mrrs4}. We know the reduced calculation is mainly in the regularization process because we only use half of the parameters. Thus we experimented where the regularization term is only half of the parameters on the original model (See results for $5^{\bigstar}\mathrm{E}_r$) to explore whether the effect of our method is similar to the reduced parameters regularization. 

Then we experimented with conjugations in different positions to explore how the models perform differently. We set negative conjugation $c = -\overline{b}, d = \overline{a}$ in model $5^{\bigstar}\mathrm{\epsilon}_n$, where half of the conjugate parameters are using negative conjugation instead of positive conjugation; we set vertical conjugation $c = \overline{a}, d = \overline{b}$ in model $5^{\bigstar}\mathrm{\epsilon}_v$, where parameters are conjugated in their vertical direction instead of the diagonal direction; and we let $b = \overline{a}, d = \overline{c}$ in model $5^{\bigstar}\mathrm{\epsilon}_h$, the horizontal conjugation, where parameters are conjugated in their horizontal direction.

The studies show that, first, by comparing the accuracy of $5^{\bigstar}\mathrm{\epsilon}$ and $5^{\bigstar}\mathrm{E}_r$, we know that reducing parameters in the regularization process hurts the accuracy significantly, which indicates our conjugation method indeedly reserves model’s ability even when the parameters are reduced. Second, the negative conjugate model $5^{\bigstar}\mathrm{\epsilon}_n$ performs as well as $5^{\bigstar}\mathrm{\epsilon}$. Last but not least, conjugate method should choose suitable positions, e.g., $5^{\bigstar}\mathrm{\epsilon}_v$ and $5^{\bigstar}\mathrm{\epsilon}_h$ do not perform as well.

\subsection{Statistical Methods}
To clarify the difference between original models and their conjugate models, we took the highest mean as the best result, with the standard deviation as a secondary judgement, and ultimately two-sample t-tests (See Table \ref{ttest} in Appendix) are conducted to decide whether two similar results can be considered statistically equivalent and which is the best.

The two-sample t-test estimates if two population means are equal. Here we use the t-test to judge if the Time or MRR means of two models are equal. We set significance level $\alpha=0.05$, and the null hypothesis assumed that the two data samples are from normal distributions with unknown and unequal variances. $(h,p)$ means the result $h$ and $p$-value of the hypothesis test. $h=1,0$. $h=1$ means rejection to the null hypothesis at the significance level $\alpha$. $h=0$ indicates the failure to reject the null hypothesis at the significance level $\alpha$. $p \in [0,1]$ is a probability of observing a test statistic as extreme as, or more extreme than, the observed value under the null hypothesis. A small $p$ value suggests suspicion on the validity of the null hypothesis.

To prepare data for the t-tests, experiments on $\mathrm{ComplEx}$, $\mathrm{Compl\epsilon x}$, $5^{\bigstar}\mathrm{E}$, $5^{\bigstar}\mathrm{\epsilon}_\nabla$, $5^{\bigstar}\mathrm{\epsilon}_\diamondsuit$ and $5^{\bigstar}\mathrm{\epsilon}_n$ are conducted 17 times each. Apart from that, the $5^{\bigstar}\mathrm{E}_r$, $5^{\bigstar}\mathrm{\epsilon}_v$, $5^{\bigstar}\mathrm{\epsilon}_h$ apparently perform worse than the former six models, thus the t-tests are not needed and their experiments are conducted 5 times each.

Most of our t-tests were done among the original model and its conjugate models as the distribution differs significantly if the base model is different. However, since the accuracies among different models are similar on the YAGO3-10 and WN18 datasets, we did several supplementary t-tests (indicated in \textit{italics}). The supplementary t-tests showed that the distributions are different indeed when based on different original models even though they appear to be similar. On the contrary, there exist similar distributions among the results distribution of the original model and its conjugate model.

\subsection{Advantages of Parameter Sharing}
Approching for the best accuracy in link prediction task has the trade off of misinformation effect or inevitable high memory and time costs. Our parameter-sharing method by using half conjugate parameters is very easy to apply and can help control these costs, and potentially no trade off.

The original $\mathrm{ComplEx}$ and $5^{\bigstar}\mathrm{E}$ each has their own strength in the perspective of accuracy on different datasets; while $\mathrm{ComplEx}$ costs much more memory and time than $5^{\bigstar}\mathrm{E}$ when compared under similar accuracy.

Our conjugate models consume less memory and time, and not inferior to the original models in accuracy, which shows that our parameter-sharing method makes a complex number represented KGE model superior to itself.

\section{Conclusions}

We propose using shared conjugate parameters for transformations, which suffices to accurately represent the structures of the KG.

Our method can help scaling up KG with less carbon footprints easily: first, it reduces parameter size and consumes less or at least comparable training time while achieving consistent accuracy as the non-conjugate model, including reaching state-of-the-art results; second, it is easily generalizable across various complex number represented models.

\section{Future Work}


We would like to deal with the interpretation of the linear constrain of our method. For example, to explore the effect of this method on different relation patterns. Moreover, many KG applications like the work done by \citet{Hon21} regard visual appeal as important, where appropriate visuals can better convey the points of the data and facilitate user interaction. We can see that the vector representations of transformed entities using this method have more substantial geometric constrains (See transformed entities illustrated in Figure \ref{entities}). We want to explore if our method can obtain better KG visualization.



\section*{Acknowledgements}
This work was supported by JSPS KAKENHI Grant Numbers 21H05054.

\bibliography{anthology,custom}

\begin{thebibliography}{14}
\expandafter\ifx\csname natexlab\endcsname\relax\def\natexlab#1{#1}\fi

\bibitem[{Balazevic et~al.(2019)Balazevic, Allen, and Hospedales}]{Bal19}
Ivana Balazevic, Carl Allen, and Timothy Hospedales. 2019.
\newblock Multi-relational poincar\'{e} graph embeddings.
\newblock In \emph{Advances in Neural Information Processing Systems},
  volume~32. Curran Associates, Inc.

\bibitem[{Bordes et~al.(2013)Bordes, Usunier, Garcia-Duran, Weston, and
  Yakhnenko}]{Bor13}
Antoine Bordes, Nicolas Usunier, Alberto Garcia-Duran, Jason Weston, and Oksana
  Yakhnenko. 2013.
\newblock Translating embeddings for modeling multi-relational data.
\newblock In \emph{Advances in Neural Information Processing Systems},
  volume~26. Curran Associates, Inc.

\bibitem[{Chami et~al.(2020)Chami, Wolf, Juan, Sala, Ravi, and R{\'e}}]{Cha20}
Ines Chami, Adva Wolf, Da-Cheng Juan, Frederic Sala, Sujith Ravi, and
  Christopher R{\'e}. 2020.
\newblock \href {https://doi.org/10.18653/v1/2020.acl-main.617}
  {Low-dimensional hyperbolic knowledge graph embeddings}.
\newblock In \emph{Proceedings of the 58th Annual Meeting of the Association
  for Computational Linguistics}, pages 6901--6914, Online. Association for
  Computational Linguistics.

\bibitem[{Dettmers et~al.(2018)Dettmers, Minervini, Stenetorp, and
  Riedel}]{Det18}
Tim Dettmers, Pasquale Minervini, Pontus Stenetorp, and Sebastian Riedel. 2018.
\newblock Convolutional 2d knowledge graph embeddings.
\newblock In \emph{AAAI}.

\bibitem[{Hayashi and Shimbo(2017)}]{hayashi-shimbo-2017-equivalence}
Katsuhiko Hayashi and Masashi Shimbo. 2017.
\newblock \href {https://doi.org/10.18653/v1/P17-2088} {On the equivalence of
  holographic and complex embeddings for link prediction}.
\newblock In \emph{Proceedings of the 55th Annual Meeting of the Association
  for Computational Linguistics (Volume 2: Short Papers)}, pages 554--559,
  Vancouver, Canada. Association for Computational Linguistics.

\bibitem[{Hongwimol et~al.(2021)Hongwimol, Kehasukcharoen, Laohawarutchai,
  Lertvittayakumjorn, Ng, Lai, Liu, and Vateekul}]{Hon21}
Pollawat Hongwimol, Peeranuth Kehasukcharoen, Pasit Laohawarutchai, Piyawat
  Lertvittayakumjorn, Aik~Beng Ng, Zhangsheng Lai, Timothy Liu, and Peerapon
  Vateekul. 2021.
\newblock \href {https://doi.org/10.18653/v1/2021.acl-demo.14} {{ESRA}:
  Explainable scientific research assistant}.
\newblock In \emph{Proceedings of the 59th Annual Meeting of the Association
  for Computational Linguistics and the 11th International Joint Conference on
  Natural Language Processing: System Demonstrations}, pages 114--121, Online.
  Association for Computational Linguistics.

\bibitem[{Ji et~al.(2015)Ji, He, Xu, Liu, and Zhao}]{Jie15}
Guoliang Ji, Shizhu He, Liheng Xu, Kang Liu, and Jun Zhao. 2015.
\newblock \href {https://doi.org/10.3115/v1/P15-1067} {Knowledge graph
  embedding via dynamic mapping matrix}.
\newblock In \emph{Proceedings of the 53rd Annual Meeting of the Association
  for Computational Linguistics and the 7th International Joint Conference on
  Natural Language Processing (Volume 1: Long Papers)}, pages 687--696,
  Beijing, China. Association for Computational Linguistics.

\bibitem[{Lin et~al.(2015)Lin, Liu, Sun, Liu, and Zhu}]{Lin15}
Yankai Lin, Zhiyuan Liu, Maosong Sun, Yang Liu, and Xuan Zhu. 2015.
\newblock Learning entity and relation embeddings for knowledge graph
  completion.
\newblock \emph{Proceedings of the AAAI Conference on Artificial Intelligence},
  29(1).

\bibitem[{Nayyeri et~al.(2021)Nayyeri, Vahdati, Aykul, and Lehmann}]{Nay21}
Mojtaba Nayyeri, Sahar Vahdati, Can Aykul, and Jens Lehmann. 2021.
\newblock 5* knowledge graph embeddings with projective transformations.
\newblock \emph{Proceedings of the AAAI Conference on Artificial Intelligence},
  35(10):9064--9072.

\bibitem[{Nickel et~al.(2011)Nickel, Tresp, and Kriegel}]{Nic11}
Maximilian Nickel, Volker Tresp, and Hans-Peter Kriegel. 2011.
\newblock A three-way model for collective learning on multi-relational data.
\newblock In \emph{ICML}.

\bibitem[{Noy et~al.(2019)Noy, Gao, Jain, Narayanan, Patterson, and
  Taylor}]{industry-kg}
Natasha Noy, Yuqing Gao, Anshu Jain, Anant Narayanan, Alan Patterson, and Jamie
  Taylor. 2019.
\newblock \href
  {https://cacm.acm.org/magazines/2019/8/238342-industry-scale-knowledge-graphs/fulltext}
  {Industry-scale knowledge graphs: Lessons and challenges}.
\newblock \emph{Communications of the ACM}, 62 (8):36--43.

\bibitem[{Toutanova and Chen(2015)}]{Tou15}
Kristina Toutanova and Danqi Chen. 2015.
\newblock \href {https://doi.org/10.18653/v1/W15-4007} {Observed versus latent
  features for knowledge base and text inference}.
\newblock In \emph{Proceedings of the 3rd Workshop on Continuous Vector Space
  Models and their Compositionality}, pages 57--66, Beijing, China. Association
  for Computational Linguistics.

\bibitem[{Trouillon et~al.(2016)Trouillon, Welbl, Riedel, Gaussier, and
  Bouchard}]{Tro16}
Th{\'{e}}o Trouillon, Johannes Welbl, Sebastian Riedel, {\'{E}}ric Gaussier,
  and Guillaume Bouchard. 2016.
\newblock \href {http://arxiv.org/abs/1606.06357} {Complex embeddings for
  simple link prediction}.
\newblock \emph{CoRR}, abs/1606.06357.

\bibitem[{Yang et~al.(2014)Yang, Yih, He, Gao, and Deng}]{Yan15}
Bishan Yang, Wen-tau Yih, Xiaodong He, Jianfeng Gao, and Li~Deng. 2014.
\newblock \href {https://doi.org/10.48550/ARXIV.1412.6575} {Embedding entities
  and relations for learning and inference in knowledge bases}.

\end{thebibliography}
\bibliographystyle{acl_natbib}

\clearpage
\appendix
\begin{table*}														
\section{Two-sample t-test}														
\begin{subtable}[htbp]{1\textwidth}														
\centering														
\resizebox{1\textwidth}{!}{														
\begin{tabular}{lllllll}														
	& \multicolumn{6}{c}{Time t-test (h, p)}\\													
	\cline{2-7}													
		&	$\mathrm{ComplEx}$	&	$\mathrm{Compl\epsilon x}$	&	$5^{\bigstar}\mathrm{E}$	&	$5^{\bigstar}\mathrm{\epsilon}_\nabla$	&	$5^{\bigstar}\mathrm{\epsilon}_\diamondsuit$	&	$5^{\bigstar}\mathrm{\epsilon}_n$	\\
	$\mathrm{ComplEx}$	&	-	&	(0, 2e-1)	&	-	&	-	&	-	&	-	\\
	$\mathrm{Compl\epsilon x}$	&	-	&	-	&	-	&	-	&	-	&	-	\\
	$5^{\bigstar}\mathrm{E}$	&	-	&	-	&	-	&	(1, 8e-3)	&	(0, 6e-1)	&	(1, 4e-5)	\\
	$5^{\bigstar}\mathrm{\epsilon}_\nabla$	&	-	&	-	&	-	&	-	&	(0, 4e-1)	&	(0, 4e-1)	\\
	$5^{\bigstar}\mathrm{\epsilon}_\diamondsuit$	&	-	&	-	&	-	&	-	&	-	&	(0, 2e-1)	\\
	$5^{\bigstar}\mathrm{\epsilon}_n$	&	-	&	-	&	-	&	-	&	-	&	-	\\
	\hline													
\end{tabular}														
\begin{tabular}{llllll}														
	\multicolumn{6}{c}{MRR t-test (h, p)}\\													
	\hline													
	$\mathrm{ComplEx}$	&	$\mathrm{Compl\epsilon x}$	&	$5^{\bigstar}\mathrm{E}$	&	$5^{\bigstar}\mathrm{\epsilon}_\nabla$	&	$5^{\bigstar}\mathrm{\epsilon}_\diamondsuit$	&	$5^{\bigstar}\mathrm{\epsilon}_n$	\\		
	-	&	(1, 8e-17)	&	-	&	-	&	-	&	-	\\		
	-	&	-	&	-	&	-	&	-	&	-	\\		
	-	&	-	&	-	&	(1, 1e-12)	&	(1, 1e-14)	&	(1, 1e-10)	\\		
	-	&	-	&	-	&	-	&	(1, 6e-3)	&	(1, 1e-2)	\\		
	-	&	-	&	-	&	-	&	-	&	(1, 7e-6)	\\		
	-	&	-	&	-	&	-	&	-	&	-	\\		
	\hline													
\end{tabular}}														
\caption{FB15K-237}														
\end{subtable}														
\begin{subtable}[htbp]{1\textwidth}														
\centering														
\resizebox{1\textwidth}{!}{														
\begin{tabular}{lllllll}														
	& \multicolumn{6}{c}{Time t-test (h, p)}\\													
	\cline{2-7}													
		&	$\mathrm{ComplEx}$	&	$\mathrm{Compl\epsilon x}$	&	$5^{\bigstar}\mathrm{E}$	&	$5^{\bigstar}\mathrm{\epsilon}_\nabla$	&	$5^{\bigstar}\mathrm{\epsilon}_\diamondsuit$	&	$5^{\bigstar}\mathrm{\epsilon}_n$	\\
	$\mathrm{ComplEx}$	&	-	&	(0, 6e-1)	&	-	&	-	&	-	&	-	\\
	$\mathrm{Compl\epsilon x}$	&	-	&	-	&	-	&	-	&	-	&	-	\\
	$5^{\bigstar}\mathrm{E}$	&	-	&	-	&	-	&	(1, 4e-11)	&	-	&	(1, 6e-3)	\\
	$5^{\bigstar}\mathrm{\epsilon}_\nabla$	&	-	&	-	&	-	&	-	&	-	&	(1, 2e-2)	\\
	$5^{\bigstar}\mathrm{\epsilon}_\diamondsuit$	&	-	&	-	&	-	&	-	&	-	&	-	\\
	$5^{\bigstar}\mathrm{\epsilon}_n$	&	-	&	-	&	-	&	-	&	-	&	-	\\
	\hline													
\end{tabular}														
\begin{tabular}{llllll}														
	\multicolumn{6}{c}{MRR t-test (h, p)}\\													
	\hline													
	$\mathrm{ComplEx}$	&	$\mathrm{Compl\epsilon x}$	&	$5^{\bigstar}\mathrm{E}$	&	$5^{\bigstar}\mathrm{\epsilon}_\nabla$	&	$5^{\bigstar}\mathrm{\epsilon}_\diamondsuit$	&	$5^{\bigstar}\mathrm{\epsilon}_n$	\\		
	-	&	(1, 3e-28)	&	-	&	-	&	-	&	-	\\		
	-	&	-	&	-	&	-	&	-	&	-	\\		
	-	&	-	&	-	&	(1, 9e-11)	&	-	&	(1, 2e-15)	\\		
	-	&	-	&	-	&	-	&	-	&	(1, 4e-21)	\\		
	-	&	-	&	-	&	-	&	-	&	-	\\		
	-	&	-	&	-	&	-	&	-	&	-	\\		
	\hline													
\end{tabular}}														
\caption{WN18RR}														
\end{subtable}														
\begin{subtable}[htbp]{1\textwidth}														
\centering														
\resizebox{1\textwidth}{!}{														
\begin{tabular}{lllllll}														
	& \multicolumn{6}{c}{Time t-test (h, p)}\\													
	\cline{2-7}													
		&	$\mathrm{ComplEx}$	&	$\mathrm{Compl\epsilon x}$	&	$5^{\bigstar}\mathrm{E}$	&	$5^{\bigstar}\mathrm{\epsilon}_\nabla$	&	$5^{\bigstar}\mathrm{\epsilon}_\diamondsuit$	&	$5^{\bigstar}\mathrm{\epsilon}_n$	\\
	$\mathrm{ComplEx}$	&	-	&	(0, 4e-1)	&	-	&	-	&	-	&	-	\\
	$\mathrm{Compl\epsilon x}$	&	-	&	-	&	-	&	-	&	-	&	-	\\
	$5^{\bigstar}\mathrm{E}$	&	-	&	-	&	-	&	(1, 5e-47)	&	-	&	(1, 1e-46)	\\
	$5^{\bigstar}\mathrm{\epsilon}_\nabla$	&	-	&	-	&	-	&	-	&	-	&	(0, 7e-1)	\\
	$5^{\bigstar}\mathrm{\epsilon}_\diamondsuit$	&	-	&	-	&	-	&	-	&	-	&	-	\\
	$5^{\bigstar}\mathrm{\epsilon}_n$	&	-	&	-	&	-	&	-	&	-	&	-	\\
	\hline													
\end{tabular}														
\begin{tabular}{llllll}														
	\multicolumn{6}{c}{MRR t-test (h, p)}\\													
	\hline													
	$\mathrm{ComplEx}$	&	$\mathrm{Compl\epsilon x}$	&	$5^{\bigstar}\mathrm{E}$	&	$5^{\bigstar}\mathrm{\epsilon}_\nabla$	&	$5^{\bigstar}\mathrm{\epsilon}_\diamondsuit$	&	$5^{\bigstar}\mathrm{\epsilon}_n$	\\		
	-	&	(1, 3e-6)	&	\textit{(1, 2e-7)}	&	\textit{(1, 7e-3)}	&	-	&	\textit{(1, 3e-8)}	\\		
	-	&	-	&	-	&	-	&	-	&	-	\\		
	-	&	-	&		&	(1, 5e-4)	&	-	&	(0, 6e-1)	\\		
	-	&	-	&	-	&	-	&	-	&	(1, 3e-4)	\\		
	-	&	-	&	-	&	-	&	-	&	-	\\		
	-	&	-	&	-	&	-	&	-	&	-	\\		
	\hline													
\end{tabular}}														
\caption{YAGO3-10}														
\end{subtable}														
\begin{subtable}[htbp]{1\textwidth}														
\centering														
\resizebox{1\textwidth}{!}{														
\begin{tabular}{lllllll}														
	& \multicolumn{6}{c}{Time t-test (h, p)}\\													
	\cline{2-7}													
		&	$\mathrm{ComplEx}$	&	$\mathrm{Compl\epsilon x}$	&	$5^{\bigstar}\mathrm{E}$	&	$5^{\bigstar}\mathrm{\epsilon}_\nabla$	&	$5^{\bigstar}\mathrm{\epsilon}_\diamondsuit$	&	$5^{\bigstar}\mathrm{\epsilon}_n$	\\
	$\mathrm{ComplEx}$	&	-	&	(0, 1e-1)	&	-	&	-	&	-	&	-	\\
	$\mathrm{Compl\epsilon x}$	&	-	&	-	&	-	&	-	&	-	&	-	\\
	$5^{\bigstar}\mathrm{E}$	&	-	&	-	&	-	&	(1, 2e-6)	&	(1, 3e-5)	&	(1, 8e-3)	\\
	$5^{\bigstar}\mathrm{\epsilon}_\nabla$	&	-	&	-	&	-	&	-	&	(1, 3e-2)	&	(0, 1e-1)	\\
	$5^{\bigstar}\mathrm{\epsilon}_\diamondsuit$	&	-	&	-	&	-	&	-	&	-	&	(0, 5e-1)	\\
	$5^{\bigstar}\mathrm{\epsilon}_n$	&	-	&	-	&	-	&	-	&	-	&	-	\\
	\hline													
\end{tabular}														
\begin{tabular}{llllll}														
	\multicolumn{6}{c}{MRR t-test (h, p)}\\													
	\hline													
	$\mathrm{ComplEx}$	&	$\mathrm{Compl\epsilon x}$	&	$5^{\bigstar}\mathrm{E}$	&	$5^{\bigstar}\mathrm{\epsilon}_\nabla$	&	$5^{\bigstar}\mathrm{\epsilon}_\diamondsuit$	&	$5^{\bigstar}\mathrm{\epsilon}_n$	\\		
	-	&	(0, 8e-2)	&	-	&	-	&	-	&	-	\\		
	-	&	-	&	-	&	-	&	-	&	-	\\		
	-	&	-	&	-	&	(1, 2e-21)	&	(1, 2e-2)	&	(1, 6e-21)	\\		
	-	&	-	&	-	&	-	&	(1, 1e-23)	&	(1, 3e-9)	\\		
	-	&	-	&	-	&	-	&	-	&	(1, 3e-22)	\\		
	-	&	-	&	-	&	-	&	-	&	-	\\		
	\hline													
\end{tabular}}														
\caption{FB15K}														
\end{subtable}														
\begin{subtable}[htbp]{1\textwidth}														
\centering														
\resizebox{1\textwidth}{!}{														
\begin{tabular}{lllllll}														
	& \multicolumn{6}{c}{Time t-test (h, p)}\\													
	\cline{2-7}													
		&	$\mathrm{ComplEx}$	&	$\mathrm{Compl\epsilon x}$	&	$5^{\bigstar}\mathrm{E}$	&	$5^{\bigstar}\mathrm{\epsilon}_\nabla$	&	$5^{\bigstar}\mathrm{\epsilon}_\diamondsuit$	&	$5^{\bigstar}\mathrm{\epsilon}_n$	\\
	$\mathrm{ComplEx}$	&	-	&	(0, 5e-1)	&	-	&	-	&	-	&	-	\\
	$\mathrm{Compl\epsilon x}$	&	-	&	-	&	-	&	-	&	-	&	-	\\
	$5^{\bigstar}\mathrm{E}$	&	-	&	-	&	-	&	(1, 1e-7)	&	(1, 4e-9)	&	(1, 6e-9)	\\
	$5^{\bigstar}\mathrm{\epsilon}_\nabla$	&	-	&	-	&	-	&	-	&	(0, 2e-1)	&	(0, 8e-2)	\\
	$5^{\bigstar}\mathrm{\epsilon}_\diamondsuit$	&	-	&	-	&	-	&	-	&	-	&	(0, 1e-1)	\\
	$5^{\bigstar}\mathrm{\epsilon}_n$	&	-	&	-	&	-	&	-	&	-	&	-	\\
	\hline													
\end{tabular}														
\begin{tabular}{llllll}														
	\multicolumn{6}{c}{MRR t-test (h, p)}\\													
	\hline													
	$\mathrm{ComplEx}$	&	$\mathrm{Compl\epsilon x}$	&	$5^{\bigstar}\mathrm{E}$	&	$5^{\bigstar}\mathrm{\epsilon}_\nabla$	&	$5^{\bigstar}\mathrm{\epsilon}_\diamondsuit$	&	$5^{\bigstar}\mathrm{\epsilon}_n$	\\		
	-	&	(0, 8e-2)	&	-	&	-	&	\textit{(1, 1e-17)}	&	\textit{(1, 1e-13)}	\\		
	-	&	-	&	-	&	-	&	\textit{(1, 5e-21)}	&	\textit{(1, 2e-16)}	\\		
	-	&	-	&		&	(1, 3e-14)	&	(1, 5e-5)	&	(0, 2e-1)	\\		
	-	&	-	&	-	&	-	&	(1, 4e-15)	&	(1, 5e-14)	\\		
	-	&	-	&	-	&	-	&	-	&	(1, 3e-5)	\\		
	-	&	-	&	-	&	-	&	-	&	-	\\		
	\hline													
\end{tabular}}														
\caption{WN18}														
\end{subtable}														
\caption{t-test (h, p) of Time and MRR on FB15K-237, WN18RR, YAGO3-10, FB15K and WN18 datasets.}														
\label{ttest}														
\end{table*}														

\end{document}